\documentclass[preprint,onecolumn]{elsarticle}
\usepackage[utf8]{inputenc}
\usepackage[T1]{fontenc}
\usepackage{makecell}

\usepackage{xcolor}

\usepackage[english]{babel} 

\usepackage{graphicx}

\usepackage{algorithmicx}
\usepackage{algorithm} 
\usepackage{algpseudocode} 
\usepackage{amsmath}
\usepackage{amssymb}
\usepackage{makecell} 
\usepackage{placeins}

\usepackage{url}

\begin{document}
\newpageafter{author}

\begin{frontmatter}

\author[inst1]{José María Buades Rubio } 
\ead{josemaria.buades@uib.es}
\author[inst1,inst2]{Antoni Jaume-i-Capó}
\ead{antoni.jaume@uib.es}
\author{David López González}
\ead{lopezdavidg.com@outlook.com} 
\author[inst1,inst2]{Gabriel Moyà Alcover \corref{cor}}
\ead{gabriel.moya@uib.es} 
\cortext[cor]{Corresponding author}

\affiliation[inst1]{organization={Computer Graphics and Vision and AI Group (UGiVIA), Research Institute of Health Sciences (IUNICS), Departament de Matemàtiques i Informàtica, Universitat de les Illes Balears},
            addressline={Ctra Valldemossa Km 7.5}, 
            city={Palma},
            postcode={07122}, 
            state={Illes Balears},
            country={Spain}}

\affiliation[inst2]{organization={Laboratory for Artificial Intelligence Applications (LAIA@UIB), Universitat de les Illes Balears},
             addressline={Ctra Valldemossa Km 7.5}, 
             city={Palma},
            postcode={07122}, 
             state={Illes Balears},
             country={Spain}}

\title{Solving nonograms using Neural Networks} 






\begin{abstract}
Nonograms are logic puzzles in which cells in a grid must be colored or left blank according to the numbers that are located in its headers. In this study, we analyze different techniques to solve this type of logical problem using an Heuristic Algorithm, Genetic Algorithm, and Heuristic Algorithm with Neural Network. Furthermore, we generate a public dataset to train the neural networks.  We published this dataset and the code of the algorithms. Combination of the heuristic algorithm with a neural network obtained the best results. From state of the art review, no previous works used neural network to solve nonograms, nor combined a network with other algorithms to accelerate the resolution process.
\end{abstract}

\begin{highlights}
\item We analyze and compared different techniques to solve nonograms.
\item We combined the Heuristic algorithm with neural networks.
\item We generate a public dataset of nonograms to train the neural networks.
\end{highlights}


\end{frontmatter}

\section{Introduction}\label{introduction}

A nonogram, which is also known as a Picross or Hanjie, is a Japanese logic puzzle in which cells in a grid must be colored or left blank according to a set of numbers that is located at the side of the board, also known as row and column headers, to reconstruct a binary image. Each header indicates the number of cells that must be marked in a row inside the board to construct a block. If there is more than one number in the same row or column header, at least one empty cell must exist between them. Puzzles of an arbitrary size can be defined as rectangular or square. The cells of a nonogram are defined by two states: filled ($\mid  \blacksquare \mid $) and empty ($\mid$ x $\mid$).

\begin{figure}[h!]
	\begin{center}
		\includegraphics[width = 0.75\linewidth]{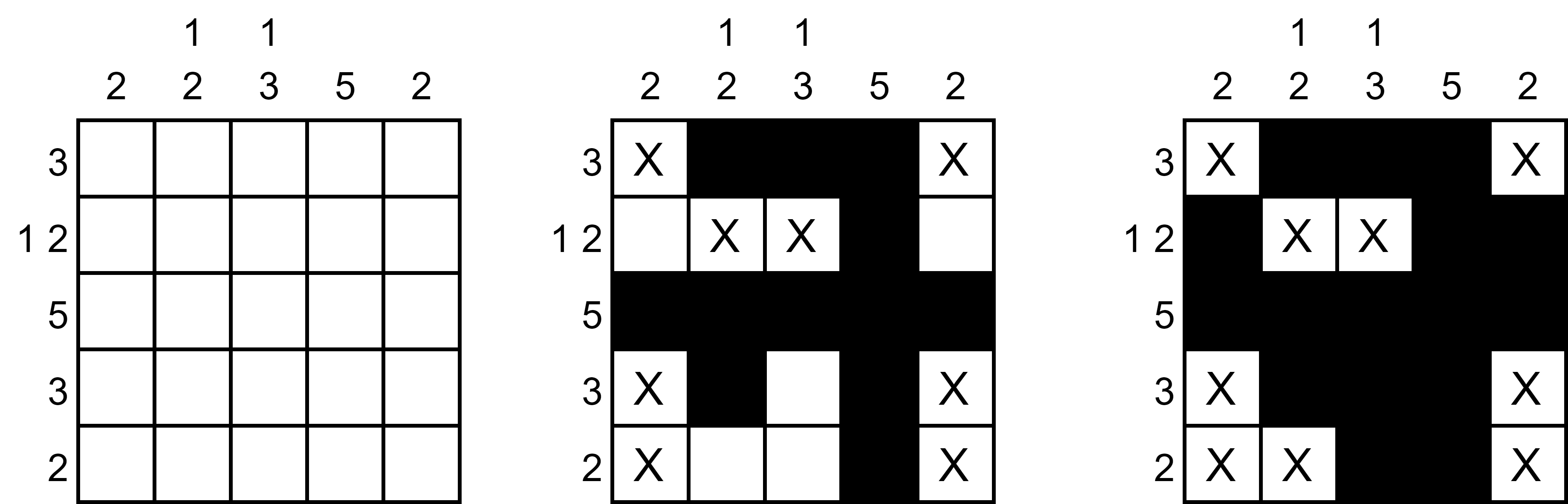}
	\end{center}
	\caption{Examples of different nonogram states: unsolved, partially solved, and solved. The black cells are considered as filled, whereas those with a cross are empty.} \label{fig:fig1}
\end{figure}

\begin{figure}[htpb]
    \centering
    \includegraphics[width = 0.60\linewidth]{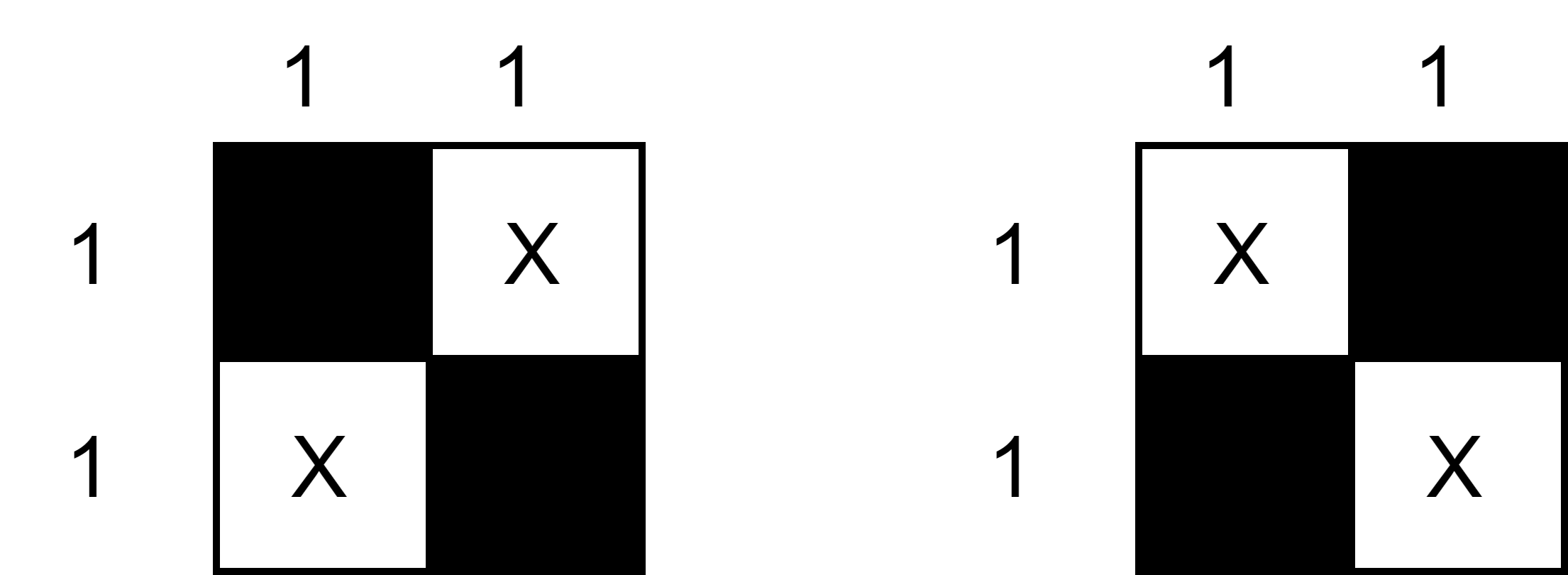}
    \caption{Nonogram with multiple solutions.} \label{fig:fig2}
\end{figure}

Figure \ref{fig:fig1} depicts the three stages of nonogram resolution:  unsolved, partially solved, and solved. Note that this type of problem falls into the category of NP completeness \cite{Ueda96np-completenessresults, Rijn2012PlayingGT, benton2006combinatorial}; thus, a solution cannot be obtained in polynomial time. Moreover, certain nonograms do not have a single solution, and all solutions that are compatible with the constraints defined by 
their headers are valid. An example of the situation is illustrated in Figure \ref{fig:fig2}.

\subsection{Problem specification}

A nonogram is a board that is defined by a matrix of size $n\times m$ with $n,m \geq 1$. As stated previously, each cell has two possible states: filled and empty, which are represented by the values $\{0, 1\}$. We define $\Omega_{n \times m}$ as the space of possible values for boards of size $n\times m$: $\Omega_{n\times m}: \{0,1\}^{n\cdot m }$. For example, boards of size $5\times 5$ have $25$ cells; consequently, there are $2^{25}$ boards of this size.

In this study, we only consider squared boards; therefore, for all nonograms, $n=m$. This implies that the rows and columns of these squared boards have the same possible codifications. We use $ \Omega_n $ instead of $\Omega_{n \times n}$ to simplify the notation. Given a board of size $n$, we define $C_n$ as the space of possible headers, and denote the encoding space of the boards as $ \lvert \Phi \lvert$, where ${\lvert C_n \lvert}^{2n}$.

\begin{figure}[th]
	\begin{center}
		\includegraphics[width =0.4\linewidth]{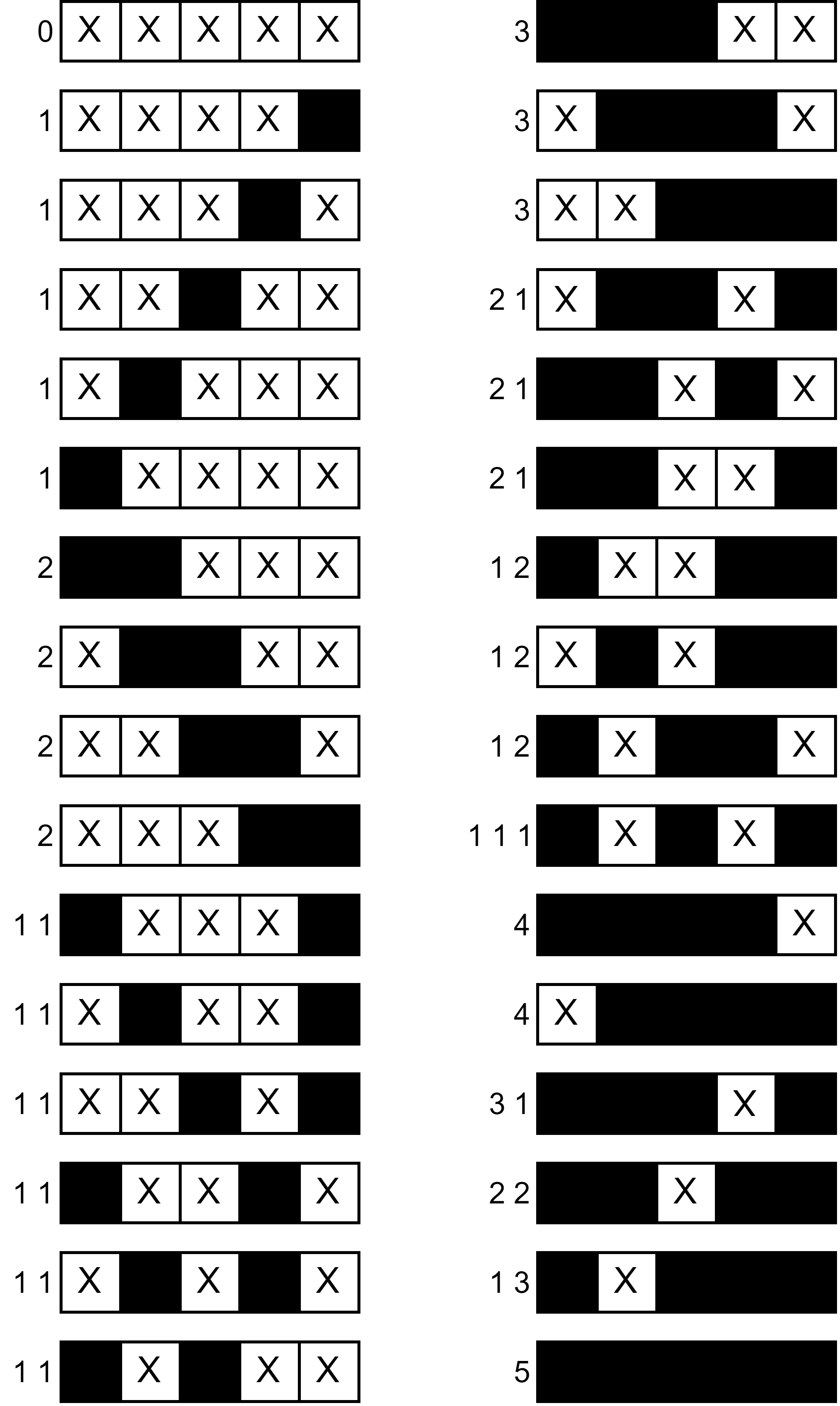}
	\end{center}
	\caption{Example of all existing configurations in a five-column nonogram. The numbers represent the row encondings.} \label{fig:fig22}
\end{figure}

If $n=5$ the set of possible header encodings is: $C_{5} = \{0, 1, 2, 3, 4, 5, 1\,1, 1\,2, \\ 2\, 1, 1\,3, 3\,1, 1\,1\,1\}$,
where $\lvert C_5 \lvert = 12$. Therefore, $\lvert \Phi_n\lvert = 12^{10}$, being $\lvert \Phi_5 \lvert > \lvert \Omega_5 \lvert$. This is because not all elements of $\Phi_5$ are valid, as they do not represent an element of $\Omega_5$, this happens for any $n$. Having $ \lvert \Phi_n \lvert > \lvert \Omega_n \lvert$ for all $n$, following the same example, for $n=5$ a column/row may have the codifications depicted in Figure \ref{fig:fig22}.

We denote $\Psi_n$ as the subset of elements of $\Phi_n$ such that they are represented by a possible board in $\Omega_n$. This implies that $\Psi_n$ contains only valid encodings. The encoding space $\Psi_n$ has fewer elements than $\Omega_n$ ($\lvert \Psi_n \lvert < \lvert \Omega_n \lvert$). Table \ref{tab:tabla_cardinales_omega_psi} displays the relationship between $\lvert \Omega \lvert $ and $\lvert \Psi \lvert$ for $n\in [1,5]$. \\

\begin{table}[htpb]
    \centering
    \begin{tabular}{|c|c|c|c|}
        Size ($n\times n$)  &  {$\lvert \Omega_n \lvert$} & {$\lvert \Psi_n \lvert$} & 
        Ratio \\
        \hline 
        $1 \times 1$ & 2 & 2 & 1 \\
        $2 \times 2$ & 16 & 15 & 0.938 \\ 
        $3 \times 3$ & 512 & 445 & 0.869 \\ 
        $4 \times 4$ & 65536 & 58196 & 0.888 \\ 
        $5 \times 5$ & 33554432 & 28781820 & 0.858 \\ 
    \end{tabular}
       \caption{For each square board sized 1 to 5, we show the number of possible boards ($\lvert \Omega_n \lvert$), number of different valid encodings ($\lvert \Psi_n \lvert$), and ratio $\lvert \Psi_n \lvert / \lvert\Omega_n\lvert $.}
    
    \label{tab:tabla_cardinales_omega_psi}
\end{table}

The coding of a board is an application $f$:
\[
f:\Omega_n \rightarrow \Psi_n.
\]

The solution of a nonogram involves a decoding process. As a step in this process, given an element $\psi \in \Psi_n$, determine an element $\omega \in \Omega_n$ that verifies $f(\omega)=\psi$. There is not always a single solution; for every $n$ there is an element $\psi \in \Psi_n$ with all row and column headers equal to 1, such that it has as the solution $n!$ elements of $\Omega_n$. 

A solver must perform the decoding process returning a compatible board with headers within the shortest amount of time. We measure the performance of different algorithms.

\subsection{State of the art}

In the literature, we identified five main strategies for solving nonograms efficiently: heuristic, depth-first search (DFS), genetic algorithms (GAs), and reinforcement learning (RL). 

\subsubsection{ Heuristic algorithms}
Regarding heuristic algorithms, Salcedo-Sanz \textit{et al.}~\cite{salcedo2007solving} designed a set of ad-hoc heuristics.  
In particular, they proposed a combinatorial ad-hoc heuristic based on trying feasible combinations of solutions in each row and column of the puzzle that may fail in solving large puzzle and a logic ad-hoc heuristic, that starts with a pre-processing step, where they filled trivial cells, and is based on the calculation of the feasible right-most and left-most solution for a given line, where right-most stands for the feasible solution which has the first filled square of each condition most in the right, that solved all the 19 nonograms that they used to validate their proposal in less than half a second. Batenburg and Kosters~\cite{batenburg2009solving} proposed a reasoning framework that can be used to determine the value of certain pixels in a nonogram; by iterating their procedure, starting from an empty grid, it is possible to solve the nonogram completely. Another rules-based approach was described in~\cite{khan2020solving}, and Khan implemented a new integer linear programming formulation, that is an optimization algorithm where the variables are integer values and the objective function and equations are linear.

\subsubsection{Depth-first search algorithms}
Concerning DFS, researchers~\cite{Jing2009, Yu2011, stefani2012solving} combined DFS algorithms and the use of logical rules. The key concept is to obtain as much information as possible by applying logical rules to detect cells which can be determined immediately at first and then applying backtracking to search for the puzzle solution. Yu \textit{et al.}~\cite{Yu2011} improved the search process using information that was obtained from a  two sets of logical rules, first set contain 5 rules and the second set contain 3 rules to refine the results obtained first set, then they applied a backtracking algorithm. Jing \textit{et al.}~\cite{Jing2009} proposed a solution based on the fact that most of Japanese puzzle are compact and contiguous, they defined a set of 11 rules divided in three main parts and then they applied the branch-and-bound algorithm to improve the search process. Finally, Stefani \textit{et al.}~\cite{stefani2012solving} applied rule–based techniques that consist of simple boxes, simple spaces, forcing, and contradiction then the best-first search is used to solve the puzzle.

Wikeckowski \textit{et al.}~\cite{wikeckowski2021algorithms} modified the DFS method, they also used a soft computing method based on permutation generation and used both to solve nonograms. The two proposed algorithms were analyzed to obtain a solution, the number of iterations, and the time required to obtain the final state. They concluded that, in contrast to the soft computing method, the DFS algorithm guaranteed correct solutions regardless of the level of difficulty of the nonogram.

\subsubsection{ Genetic algorithms}
Respecting GA, Tsai~\cite{tsai2012solvings} proposed a Taguchi-based GA that was effectively applied to solve nonograms, and tested their approach on large nonograms (from $15\times15$ to $30\times25$ ), managing to solve more than 50\% of the nonograms. Bobko and Grzywacz \cite{bobko2016solving} presented the concept of using classical GAs as a tool for solving nonograms. They obtained good results with nonograms of sizes $4\times4$ and $5\times5$, but in the case of $10\times10$, with a chromosome length of 100, none of their proposals yielded the expected result. Soto \textit{et al.}~\cite{Soto2016} developed a GA  that uses a new generation of the initial population and experimentally applied it to $5\times5$ boards. 

Habes and Hasan~\cite{Alkhraisat2016} proposed an adaptation of the particle filter, called Particle Swarm Optimization which is a population based stochastic optimization method, to solve nonograms and performed an experiment using only three boards with different sizes. 

{ Chen and Lin \cite{chen2019fast} proposed to add a new parameter in the dynamic programming approach to solve nonograms to finish maximal painting earlier without significant overhead expense. Their approach was designed to solve nonograms of size 5x5 under tournament constraints.}

\subsubsection{Reinforcement learning algorithms}
Finally,  as for RL ,Shultz \textit{et al.}~\cite{Thomas2012} applied RL to solve this problem. They found that the RL solvers learned near-optimal solutions that outperformed a heuristic solver based on general rules, but they only applied their solution to solve $5\times5$ boards because of the computational time required for larger puzzles.

\subsection{ Research objective}
Our main objective was to determine whether a neural network can improve the performance of two well-known algorithms (DFS and GA) to solve nonograms. The hypothesis was neural networks can help to solve nonogram efficiently From literature review, no previous works used a neural network to solve nonograms, nor combined a network with other algorithms to accelerate the resolution process. For this reason, in this study, we developed two new approaches. First, we used a combination of the DFS and a neural network. Second, we combined a GA and a neural network. Furthermore, we developed a dataset to evaluate the proposed solutions. This dataset and the algorithm code were published.  %
\subsection{Outline}

The remainder of this paper is organized as follows: In Section (\ref{sec:methods}), we explain the methods that we developed to solve nonograms. Section (\ref{sec:exp}) presents the experimental setup, including a description of the dataset that we created, the metrics used, and the experiments that were performed. In Section (\ref{sec:resultados}), we outline the obtained results and discuss the performance of the proposed methods. Finally, Section (\ref{sec:conclusion}) summarizes the conclusions and presents the strengths and shortcomings of the proposed system.

\section{Methods}\label{sec:methods}

In this section, we explain the networks that we designed, as well as the algorithms that we used for solving the boards. These well-known algorithms were modified to include the networks as a support for the decision-making process.

\subsection{Network design} \label{sec:methods:deeplearning}

\begin{figure*}
	\begin{center}
		\includegraphics[width = \linewidth]{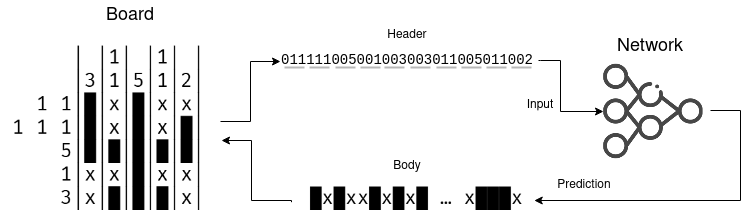}
	\end{center}
	\caption{Given the nonogram on the left, we codify its headers and construct a one-dimensional vector. This vector is the network input. As a result, we obtain a vector of size $width\cdot height$ that we can easily transform into a nonogram board.} \label{fig:fig23}
\end{figure*}

We employed a fully connected (FC) neural network, which is an artificial neural network that is composed solely of fully connected layers. The input to the network was a vector of size $LM\cdot width+LM\cdot height$, where $width= height= n$; $LM = \lfloor (n+1)/2 \rfloor$, and $n$ was the board size. We constructed this vector by concatenating all headers. To make them equal in size, we added leading zeros to each header until its size was equal to $LM$ (see Figure \ref{fig:fig23} for details). 

The output of the network was a vector of size $width\cdot height$ and each element of this vector had two possible states: empty or filled. This vector was converted into a board shape of size $(width, height)$ by applying a simple transformation. See Figure \ref{fig:fig23} for further details.

Owing to the characteristics of the designed networks, they require an input of a fixed size; thus, a model that is trained for a $5\times5$ board will not be useful for solving $10\times10$ or $15\times15$ boards.

\subsection{Reflections}

As stated previously, we can represent nonograms in two spaces: the board space $\Omega_n$ and header space $\Psi_n$. The space $\Omega$ represents all existing boards. For example, for a $5\times5$ board using two states, there are $2^{25}$  possibilities. The space $\Psi_n$ represents all valid header combinations; in the case of a $5\times5$ board, the number is $28781820$. Moreover, several elements of the space $\Psi$ can represent different boards; that is, the encoding operation is not bijective, as illustrated in Figure \ref{fig:fig2}.  

To improve the network performance, we searched for bijective functions in the $\Psi$ space, which implied a bijection in the $\Omega$ space. Reflections are functions that fulfill this requirement. By applying these reflections, we could improve the network results; if the network learned one of the reflections, it could make a correct prediction.  
We defined eight bijective functions that form a noncommutative group, which can be generated using the following three functions and identity ($id$): swapping rows and columns ($g$: diagonal reflection, where $g\circ g = id$), reversing the row order ($f_v$: vertical reflection, where $f_v\circ f_v = id$), and reversing the column order ($f_h$: horizontal reflection, where $f_h\circ f_h = id$). 

\begin{figure}[t!]
	\begin{center}
		\includegraphics[width = 0.4\linewidth]{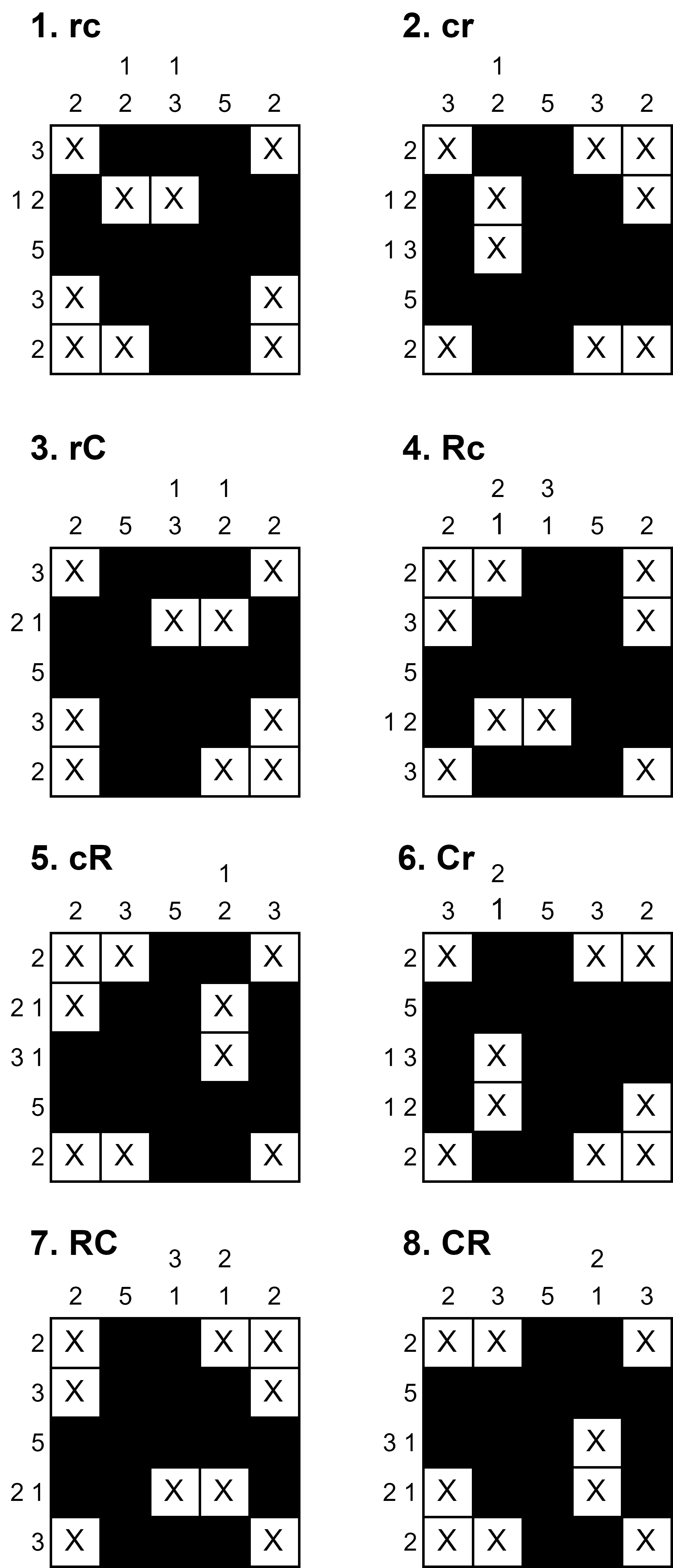}
	\end{center}
	\caption{Visual description of the reflections on a $5\times5$ board.} \label{fig:fig10}
\end{figure}

The vertical and horizontal reflections produce header realignment and provoke a swap of the header values. For example, if a row has header \textit{1 2} when horizontal reflection is applied, it becomes \textit{2 1}. The eight bijective functions are depicted in Figure \ref{fig:fig10} and explained below.

\begin{enumerate} 

\item \textbf{rc} $\rightarrow$ Rows first, columns second. Identity function $id$.

\item \textbf{cr} $\rightarrow$ Columns first, rows second. Diagonal reflection $g$. Interchange rows and columns.

\item \textbf{rC} $\rightarrow$ Rows first, then columns in reverse order. Horizontal reflection $f_h$. Invert column order and invert row numbers. 

\item \textbf{Rc} $\rightarrow$ Rows first in reverse order, then columns. Vertical reflection $f_v$. Invert row order and invert row numbers.

\item \textbf{cR} $\rightarrow$ Columns first, then rows in reverse order. Composed of $f_h \circ g = g \circ f_v$. 

\item \textbf{Cr} $\rightarrow$ Columns first in reverse order, then rows. Composed of $f_v \circ g = g \circ f_h$.

\item \textbf{RC} $\rightarrow$ Rows first in reverse order,  then columns in reverse order. Composed of horizontal and vertical reflection $f_v \circ f_h = f_h \circ f_v$. 

\item \textbf{CR} $\rightarrow$ Columns first in reverse order, then rows in reverse order. Composed of $f_v \circ f_h \circ g= f_h \circ f_v \circ g$.

\end{enumerate}

\subsection{Algorithms} \label{sec:methods:algorithms}

In this section, we describe the different approaches that we designed and implemented to solve nonograms. The first is a heuristic algorithm that we used as a baseline. We implemented a modification to introduce the neural network, as described in Section \ref{sec:methods:deeplearning}, as a support for the decision-making process and the reflections as an improvement. Furthermore, we introduced the concept of intuition, which means weighting the solution path by the similarity of the state of the board with the neural network proposal. We also describe the partial and full erase ideas that are used to accelerate the solution process. Finally, we propose a GA that also uses a neural network with the same objective, which is used in the state of the art. We were interested in evaluating their performance when we added a neural network to the decision process.

\subsubsection{Heuristic algorithm (H)}

The first algorithm is composed of heuristic rules and a DFS. The heuristic is based on the concept of trivial cells, which can be filled using the information that is available from the puzzle headers and previous board state. We used the set rules found in~\cite{Yu2011} to design this algorithm.

\begin{figure}[!ht]
	\begin{center}
		\includegraphics[width = \linewidth]{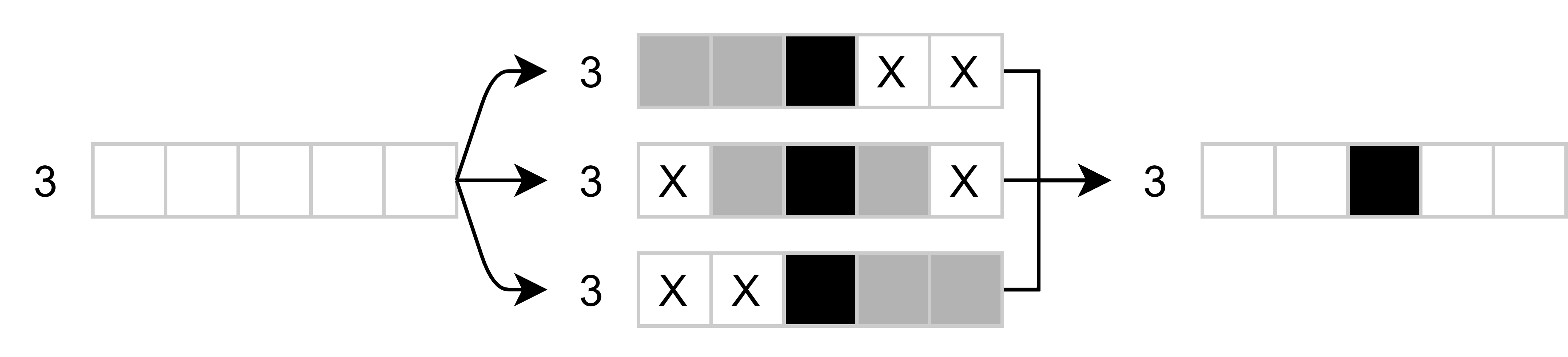}
	\end{center}
	\caption{Example of the process that we followed to determine a trivial cell. To solve a row with header $3$, there are three possible solutions, and the cell in the center is marked in all of them. We can conclude that this cell must be filled.} 
	\label{fig:fig6}
\end{figure}

Given a row or column, we establish the trivial cells by calculating all possible combinations given the header information. Subsequently, we obtain combinations that are compatible with the state of the row or column. At each step, we analyze the combinations that are compatible with the header to determine if there is a cell with the same value in all configurations to update the cell value. An example of this situation is shown in Figure \ref{fig:fig6}. This process is repeated until the nonogram is solved or until it is not possible to obtain further trivial cells.

Once we fill all of the trivial cells, the second process, namely DFS, is implemented, whereby the path selection is determined by the number of possible combinations of the row or column in ascending order. At each search step, we repeat the trivial cell-filling process before performing a new search. The heuristic algorithm is summarized in algorithm~\ref{alg:solver} and code is available at \url{https://gitlab.com/DLG-05/musi-tfm-nonograma}.

\begin{algorithm}[t!]
     \footnotesize
	\caption{Heuristic (H)}\label{alg:solver}
	\begin{algorithmic}[1]
		\Function{Solve}{board}
		\If{$Is\_Full$(board) and $Is\_Solution$(board)} 
			\State\Return True
		\ElsIf{$Is\_Full$(board)} 
			\State\Return False
		\EndIf
		\State{comb,\ trivial\_cells $\gets get\_combinations$(board)}
		\If{trivial\_cells $\not= \emptyset$ }
			\State {$t2 \gets Fill\_trivial\_cells$(board)}
			\State\Return $Solve$(t2)
		\Else
			\For {c $in \textrm{ } Sorted$(comb)}
				\State{$t2 \gets Apply\_combination$(c, board)}
				\If{$Solve$(t2)}
				    \State\Return True
				\EndIf  
			\EndFor
		\EndIf
		\State\Return False
		\EndFunction
	\end{algorithmic}
\end{algorithm}

\subsubsection{Network, heuristic algorithm and intuition (NeHI)}\label{sec:NeHI}

We modify the heuristic algorithm to integrate the neural network. This algorithm has two steps: First, we predict the board with the neural network and check if a correct solution is obtained

Second, if a solution is not obtained in the previous step, we start the DFS search using \textit{intuition} with an empty board. This concept is applied at each step of the search, and we weight each path according to the similarity to the neural network prediction. This is the number of differences between the path and network prediction. The algorithm visits them according to their scores in ascending order; in the case of a tie, we select a path randomly.

These modifications are summarized in Algorithms \ref{alg:heuristico_con_ia_1} and \ref{alg:heuristico_con_ia_2},  and code is available at \url{https://gitlab.com/DLG-05/musi-tfm-nonograma}. 

\begin{algorithm}
    \footnotesize
	\caption{NeHI - Part 1}\label{alg:heuristico_con_ia_1}
	\begin{algorithmic}[1]
		\Function{main}{{board}}
			\State $prediction \gets Get\_network\_prediction$(board)
			\If{$Is\_Solution$(prediction)} 
				\State\Return True 
			\Else
				\State \Return $Solve\_AI$(board)
				
			\EndIf
		\EndFunction
	\end{algorithmic}
\end{algorithm}

\begin{algorithm}
    \footnotesize
	\caption{NeHI - Part 2}\label{alg:heuristico_con_ia_2}
	\begin{algorithmic}[1]
		\Function{Solve\_AI}{board}
		    \If{$Is\_Full$(board) and $Is\_Solution$(board)} 
			\State\Return True
		    \ElsIf{$Is\_Full$(board)} 
			\State\Return False
		    \EndIf
			\State{comb,\ trivial\_cells $\gets get\_combinations$(board)}
	    	\If{trivial\_cells $\not= \emptyset$ }
			    \State {$t2 \gets Fill\_trivial\_cells$(board)}
			    \State\Return $Solve\_AI$(t2)
		    \Else
				\For {c $in \textrm{ } Weighted$(comb)}
				    \State{$t2 \gets Apply\_combination$(c, board)}
				    \If{$Solve\_AI$(t2)}
				        \State\Return True
				    \EndIf  
			    \EndFor
			\EndIf
		\EndFunction
	\end{algorithmic}
\end{algorithm}
 
\subsubsection{Heuristic algorithm using a network, with partial erase, full erase and intuition (NeHPF/NeHPFI)}\label{sec:HNePFI}

We designed this third version of the heuristic algorithm to improve the results of the previous algorithm when network prediction is not a valid solution. 

Following the same procedure as that described in Section \ref{sec:NeHI}, we start by predicting the nonogram using a neural network. Then, if we do not obtain a valid solution, we analyze the network result to find the rows and columns that are not correctly predicted and those that are not consistent with the header, and we erase these. This process is known as partial erase. We then start the heuristic algorithm using the board state after the partial erase. The intuition is a hyperparameter of this algorithm. 

Following the previous process, if we do not find a correct solution, we start the heuristic algorithm with an empty board without using the network. We refer to this process as full erase. 

Using this new version of the algorithm, we accelerate the solution process for most nonograms. If the board that is predicted by the network contains few errors, the algorithm can reach a final state (solution or not) in a small number of iterations, thereby avoiding the performance of the entire heuristic process to obtain a valid solution. The pseudo-code of the algorithm is presented in \ref{alg:heuristico_con_borrado_1}  and code is available at \url{https://gitlab.com/DLG-05/musi-tfm-nonograma}.

\begin{algorithm}
    \footnotesize
	\caption{NeHPF}\label{alg:heuristico_con_borrado_1}
	\begin{algorithmic}[1]
		\Function{main}{board, intuition}
		\State $prediction \gets Get\_prediction$(board)
		\If{$Is\_Solution$(prediction)} 
			\State\Return True
		\ElsIf{intuition}
			\State $t2 \gets Delete\_wrong\_rows\_columns$(prediction)
			\If{$ algorithm\_AI$(t2)} 
				\State \Return True
			\Else 
				\State \Return$Solve\_AI$(board)
			\EndIf
		\Else
			\State \Return $Solve$(board)
		\EndIf
		\EndFunction
	\end{algorithmic}
\end{algorithm}
  
\subsubsection{ Heuristic algorithm using a network trained with all reflections \\ (Ne8HI/Ne8HPF/Ne8HPFI)}
 
Based on the preliminary experiments, the network could not solve all boards, but it could solve one of its eight reflections. We design this algorithm to improve the network prediction results. 

This version includes the same algorithms as those in Sections \ref{sec:NeHI} and \ref{sec:HNePFI}. However, the networks are trained with the eight possible reflections of each nonogram of the training set, and if the network can solve any of the eight reflections, it can solve the proposed nonogram. Once one of the reflections is solved, the original board can be obtained by reversing it. It is important to note that the eight reflections are bijective functions in the header and board spaces. Code available at \url{https://gitlab.com/DLG-05/musi-tfm-nonograma}.

\subsubsection{Genetic Algorithm (GA)}

Finally, we develop a GA to solve nonograms. The mutations consist of filling, emptying, or moving filled cells. A cell is filled if the number of filled cells in the current nonogram is smaller than that in the target nonogram. We empty a cell if there were more filled cells than those in the target nonogram. Finally, if the number of filled cells is equal to that in the target nonogram, the cell is moved. The number of mutations is determined randomly between 0 and 4, and the locations are randomly selected following a uniform distribution. 

We generate the initial population of this algorithm by predicting a board using a neural network, and then we apply mutations to it. We define the following fitness function to evaluate each board: 

\begin{equation} 
   Fitness  = ncc + ncr + (width \cdot ~height) \cdot \frac{nmc}{ncmb}
\end{equation}

where \textit{ncc} denotes the number of correct columns, \textit{ncr} denotes the number of correct rows, \textit{nmc} denotes the number of marked cells, and \textit{ncmb} denotes the number of cells that must be marked to complete the board. The fitness function is based on the concepts presented in \cite{Soto2016}. 

The pseudo-code of the algorithm is presented in \ref{alg:gen}  and code is available at \url{https://gitlab.com/DLG-05/musi-tfm-nonograma}. 

\begin{algorithm}
\footnotesize
	\caption{GA}\label{alg:gen}
	
	\begin{algorithmic}[1]
		\Function{genetic}{board}
		\While{True}
			\State $Total\_weight \gets get\_total\_weight()$
			\For{i in range(len(individuals))}
				\State$ w \gets random\_uniform(0,1) \cdot Total\_weight$
				\State$ x \gets get\_individuals(w)$
				\State$ y \gets get\_mutation(x)$
				\If{$Is\_Solution$(y)}
					\State\Return True
				\EndIf
			\EndFor
		\State{$change\_population()$}
		\EndWhile
		\EndFunction
	\end{algorithmic}
\end{algorithm}

\section{Experimental setup}\label{sec:exp}
In this section we describe the dataset, the experimental environment, the experiments and the metrics. 

\subsection{Dataset}
To train the neural networks, we required a large number of nonograms with their corresponding solutions. To the best of our knowledge, there is no public dataset that fulfills our needs; therefore, we generated our own dataset. 

\begin{figure}[t] 
    \begin{center} 
        \includegraphics[width = 0.6\linewidth]{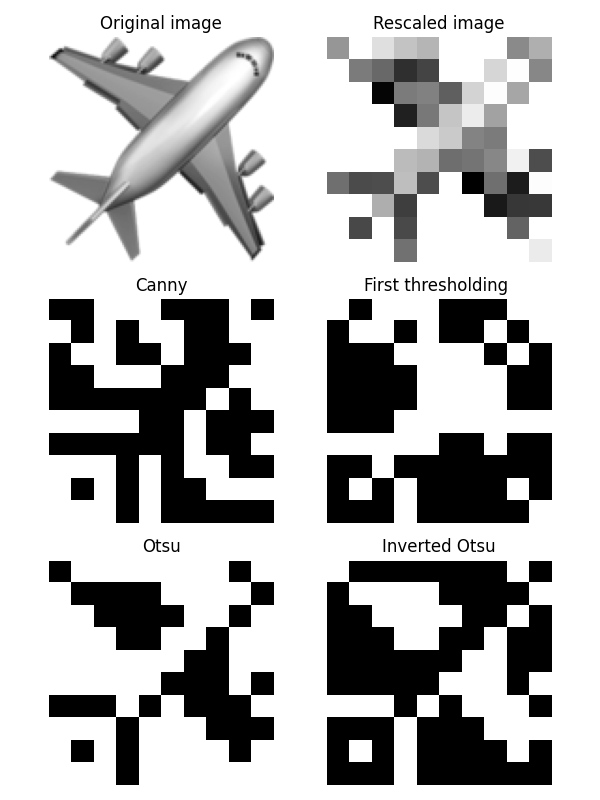} 
\end{center} 
\caption{Examples of the different methods that we used to transform an image into a nonogram. The white color denotes a filled cell.} \label{fig:fig4} 
\end{figure}

The dataset contained samples of boards of sizes $5\times5$, $10\times10$ and $15\times15$. We generated all possible boards of size $5\times5$; that is, $2^{5\times5} \simeq 3.36\mathrm{e}07$ boards. It was not possible to generate all boards of sizes $10\times10$ and $15\times15$, because the number thereof was very high: $2^{10\times10} \simeq 1.27\mathrm{e}30$ and $2^{15\times15} \simeq 5.39\mathrm{e}67$ boards. For these sizes, we employed the following strategy: First, we used the MAME Icons dataset~\cite{MAME} and Icons50 dataset~\cite{hendrycks2019robustness}, and applied the average image hash~\cite{Krawetz2011} to detect and remove duplicated images efficiently. Second, we scaled the images to the desired board size ($10\times10$ and $15\times15$ pixels). Third, we applied the following four transformations to generate the first set of nonograms: a Canny filter to detect edges, a binary threshold with a threshold value equal to 128, Otsu's threshold, and an inverted Otsu's threshold of the original image. An example of this transformation is depicted in Figure~\ref{fig:fig4}. After applying this process, a set of 76,368 boards was available for each size.

To complete the dataset, we generated a set of random nonograms because during several preliminary tests, we found that the size of the dataset was not sufficiently large to train the neural networks. We used two different techniques: we applied uniform noise between 0 and 1 to an empty board with a threshold of $0.5$, and we drew a random number of simple geometric figures (lines, rectangles, and circles) with random sizes and positions on an image. One-third of this second dataset was generated by uniform noise and the remaining two-thirds by drawing simple figures. The dataset is available at: \url{https://github.com/josebambu/NonoDataset}. 

Finally, our dataset was composed of $33,554,432$ samples of boards of size $5\times5$, and after joining the two sets, we obtained a large dataset that was composed of $76,368$ boards generated from images and $300,000$ random boards of size $10\times10$, and $76,368$ boards generated from images and $600,000$ random boards of size $15\times15$.  

\subsection{Experimental environment}
We used TensorFlow and Keras on different machines with a NVIDIA K40, NVIDIA RTX2060, and NVIDIA RTX3090 to train the neural network models.

\subsection{Experiments}

\subsubsection{Experiment 1: Neural network training} \label{sec:exp:dnn}
The first part of the experiment consisted of evaluating the performance of the different neural networks that we designed using the dataset that we described previously. The results of this experiment were used to determine the best networks, which we used in the subsequent algorithms. 

To perform this experiment, we designed a set of FC networks using three to eight hidden layers. The activation function for the intermediate layers was a ReLU and the function for the output layer was a sigmoid. We used the  Adam \cite{kingma2014adam} optimizer with different learning rates.  

To train the neural networks, we used all boards of size $5\times5$, and we used the same dataset to evaluate the network performance. We performed this simple experiment to verify the capabilities of the designed networks. To create the training and test sets for the boards of sizes $10\times10$ and $15\times15$, we divided the data generated from the images into $80\%$ for the training set and $20\%$ for the test set. Subsequently, we added all nonograms generated by the random process to the training dataset. 

Eventually, we had $33,554,432$ boards for the training and test sets of boards of size $5\times5$; $361,094$ samples for the training set and $15,274$ samples for the test set of boards of size $10\times10$; and $661,094$ boards for the training set and $15,274$ for the test set of boards of size $15\times15$. 

\subsubsection{Experiment 2: Heuristic algorithms} \label{sec:exper:algorithms2}
We designed this experiment to evaluate the heuristic algorithm, which we used as a baseline, and its three variations that used a neural network. We tested them with and without reflections, and when enabling and disabling the intuition mechanism. Table \ref{tab:table1} summarizes the different parts of the experiment. 

For each board of size $10\times10$, we tested the algorithms described in Table \ref{tab:table1} 12 times using the validation dataset to obtain valid statistical measurements. For the boards of size $15\times15$, we used only a subset of 50 samples from the validation set.  

Using this subset, we estimated the time required to solve the entire validation dataset.  

\begin{table}[htpb] 

\label{tab:table1} 
\begin{center} 
    \begin{tabular}{ll} 

        \textbf{Experiment} & \textbf{Acronym} \\\hline	 
        Heuristic algorithm & H \\\hline 

        \makecell[l]{Network, heuristic algorithm, \\ and intuition } & NeHI \\\hline 

        \makecell[l]{Network with reflections, \\  heuristic  algorithm, and intuition }  & Ne8HI \\\hline 

        \makecell[l]{Network, heuristic algorithm \\ with partial and full erase}  & NeHPF \\\hline 

        \makecell[l]{Network with  reflections, heuristic \\ algorithm with partial and full erase }  & Ne8HPF \\\hline 

        \makecell[l]{Network, heuristic algorithm with \\ partial and full erase, and intuition}  & NeHPFI \\\hline 

        \makecell[l]{Network with  reflections, heuristic \\ algorithm  with partial and \\ full erase, and intuition}  & Ne8HPFI \\\hline 

    \end{tabular} 
    \caption{Experiments performed on the heuristic algorithm and its modifications. We tested all possible combinations of the reflections and intuition hyperparameters.} 
\end{center} 
\end{table} 

\subsubsection{Experiment 3: Genetic Algorithm} \label{sec:exper:algorithms3}

Finally, we evaluated the GA with boards of size $10\times10$ using the following set of parameters: 100 iterations and a population of 1000 individuals. We did not evaluate this algorithm with boards of size $15\times15$ because a significant amount of time was required to obtain a result.

\subsection{Metrics} 

From the early empirical results, we verified that the number of erroneous cells on the boards that were predicted by the tested neural networks followed a discrete Weibull distribution. 

Given a neural network, let $F$ be the random variable that represents the number of erroneous cells that are predicted on a board by a given network. Then, we have:
 \begin{center} 
    \begin{equation} 
    P\left\lbrace F \leq f \right\rbrace= 1 -e^{-\left( \frac{f+1}{\alpha}\right)^\beta}, 
    \end{equation} 
\end{center} 
where $\alpha$ and $\beta$ are the scale and shape parameters respectively, and $f\in \{0,1,2, ..., n\times n\}$ is the number of erroneous cells. We can compare two neural networks by comparing the parameters $\alpha$ and $\beta$ of the Weibull distribution that best fit the experimentation results.

The shape value ($\beta$) of the neural networks with the best results was less than one, which implies that its mode was 0. Thus, from the neural networks with mode 0, we selected that with the smallest scale ($\alpha$), which matched the neural network with a higher number of full correct boards (0 errors). 

We used time-based metrics: the mean, standard deviation, and median of the time to solve a nonogram to evaluate the performance of the algorithms.

\section{Results and discussion} \label{sec:resultados}

In this section, we describe and discuss the results obtained from the three experiments.  

\subsection{Experiment 1: Neural networks} \label{sec:resultados:dnn}

We correctly predicted $77.06\%$ of boards of size $5\times5$. The best model architecture was composed of three layers of sizes 2048, 1024, and 256. After each layer, we added a dropout rate of 5\%  to avoid overfitting during the training phase, the learning factor was set to $0.001$, and the error function was the binary cross-entropy as each cell can have only two values. The Weibull distribution parameters were as follows: shape ($\beta$) = 0.5091745 and scale ($\alpha$) = 0.1102024.

For boards of sizes $10\times10$ and $15\times15$, the best model architecture was composed of five dense layers of sizes 2048, 1024, 1024, 1024, and 512. After each layer, we added a \textit{dropout} of $5\%$. The learning factor was $0.0001$ and the error function was \textit{binary cross-entropy}. The Weibull parameters were as follows: shape ($\beta$) = 0.420651 and scale ($\alpha$) = 3.401227 for the $10\times10$ boards, and shape ($\beta$) = 0.6219632 and scale ($\alpha$) = 18.54031 for the $15\times15$ boards. By applying this model, we correctly predicted $27.25\%$ of boards of size $10\times10$ and $5.87\%$ of boards of size $15\times15$.

\begin{figure}[ht!] 
    \begin{center} 
    \includegraphics[width = 0.65\linewidth]{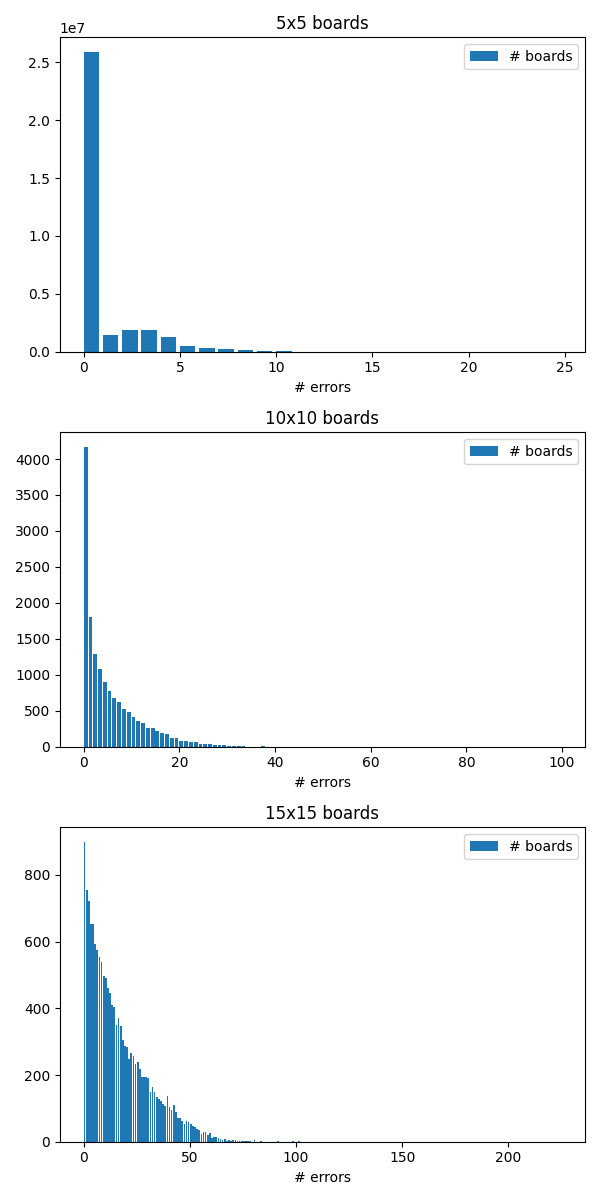} 
    \end{center} 

    \caption{Histograms of number of errors of best network for each board size. We considered an error to have occurred when a position was incorrectly predicted.} \label{fig:fig7} 
\end{figure}

The histogram of the number of errors on each board using the networks described above is depicted in Figure \ref{fig:fig7}.  

\begin{figure}[t!]
	\begin{center}
		\includegraphics[width = 0.67\linewidth]{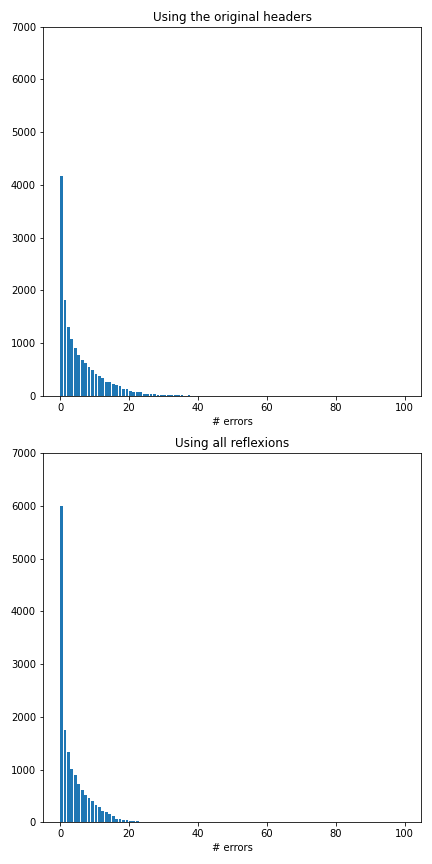}
	\end{center}
	\caption{Comparison of the performance (number of incorrectly predicted cells) of the best network for boards of size $10\times10$ when we used reflections. We can observe that the first bins in the histogram that correspond to the prediction of the networks with reflections are higher than those of the network without reflections, indicating better performance.} \label{fig:fig11}
\end{figure}

To evaluate the performance of the neural networks using reflections, we calculated the number of correct predictions of boards of size $10\times10$ when these were applied. The success rate increased from $27.25\%$ to $39.26\%$. Figure \ref{fig:fig11} indicates an improvement in the network accuracy when we used reflections, according to the histogram of the number of errors in each board.

\subsection{Experiment 2: Heuristic algorithms} \label{sec:resultados:algs}

Once we obtained the results of the neural networks, we selected the best models for each board size and evaluated the performance of the proposed algorithms using boards of size $10 \times 10$ and $15 \times 15$. 

After executing all experiments summarized in Table \ref{tab:table1} 12 times, as a first result, all algorithms solved all boards. To analyze the results in detail, we present a statistical summary of the execution time in Table \ref{tab:table2} and the number of iterations required to solve the boards in Table \ref{tab:table3}. 

\begin{table}[t!] 
	\label{tab:table2}
	\begin{center}
		\begin{tabular}{lccc}
\textbf{Algorithm} & \textbf{Mean}                             & \textbf{Std}                              & \textbf{Median}                           \\\hline
H         & 1.2003                           & 3.4236                           & 0.2042                           \\\hline
NeHI      & 1.0428                           & 3.1065                           & 0.1526                           \\\hline
Ne8HI     & 0.9811                           & 3.0475                           & 0.1003                           \\\hline
NeHPF     & 0.7936                           & 2.5936                           & 0.0858                           \\\hline
Ne8HPF    & 0.7630                           & 2.6768                           & 0.0520                           \\\hline
NeHPFI    & 0.7891                           & 2.6290                           & 0.0904                           \\\hline
Ne8HPFI   & \textbf{0.7254} & \textbf{2.5853} & \textbf{0.0519} \\\hline
\end{tabular}
\caption{Statistical measures of the execution time, in seconds, of the algorithms with boards of size $10\times10$. Algorithms that contain an 8 in their name are the same algorithms as those in the line above, but they use reflections.} 
	\end{center}
\end{table}

\begin{table}[tpb]
	
	\label{tab:table3}
	\begin{center}
		\begin{tabular}{lccc}%
		\textbf{Algorithm} & \textbf{Mean}   & \textbf{Std}     & \textbf{Median} \\ \hline
H                  & 11.0972         & 14.3556          & 6               \\\hline
NeHI               & 9.2355          & 12.8242          & 6               \\\hline
Ne8HI              & 8.5279          & 12.7462          & 6               \\\hline
NeHPF              & 7.6012          & 12.8066          & 4               \\\hline
Ne8HPF             & 6.9063          & 12.9157          & \textbf{3}      \\\hline
NeHPFI             & 7.4203          & \textbf{12.6373} & 4               \\\hline
Ne8HPFI            & \textbf{6.6978} & 12.7493          & \textbf{3}  \\\hline
		\end{tabular}
  \caption{Statistical measures of the number of iterations required to solve boards of size $10\times10$. Algorithms that contain an 8 in their name are the same algorithms as those in the line above, but they use reflections.}
	\end{center}
\end{table}

\begin{figure}[t!]
	\begin{center}
		\includegraphics[width = 0.69\linewidth]{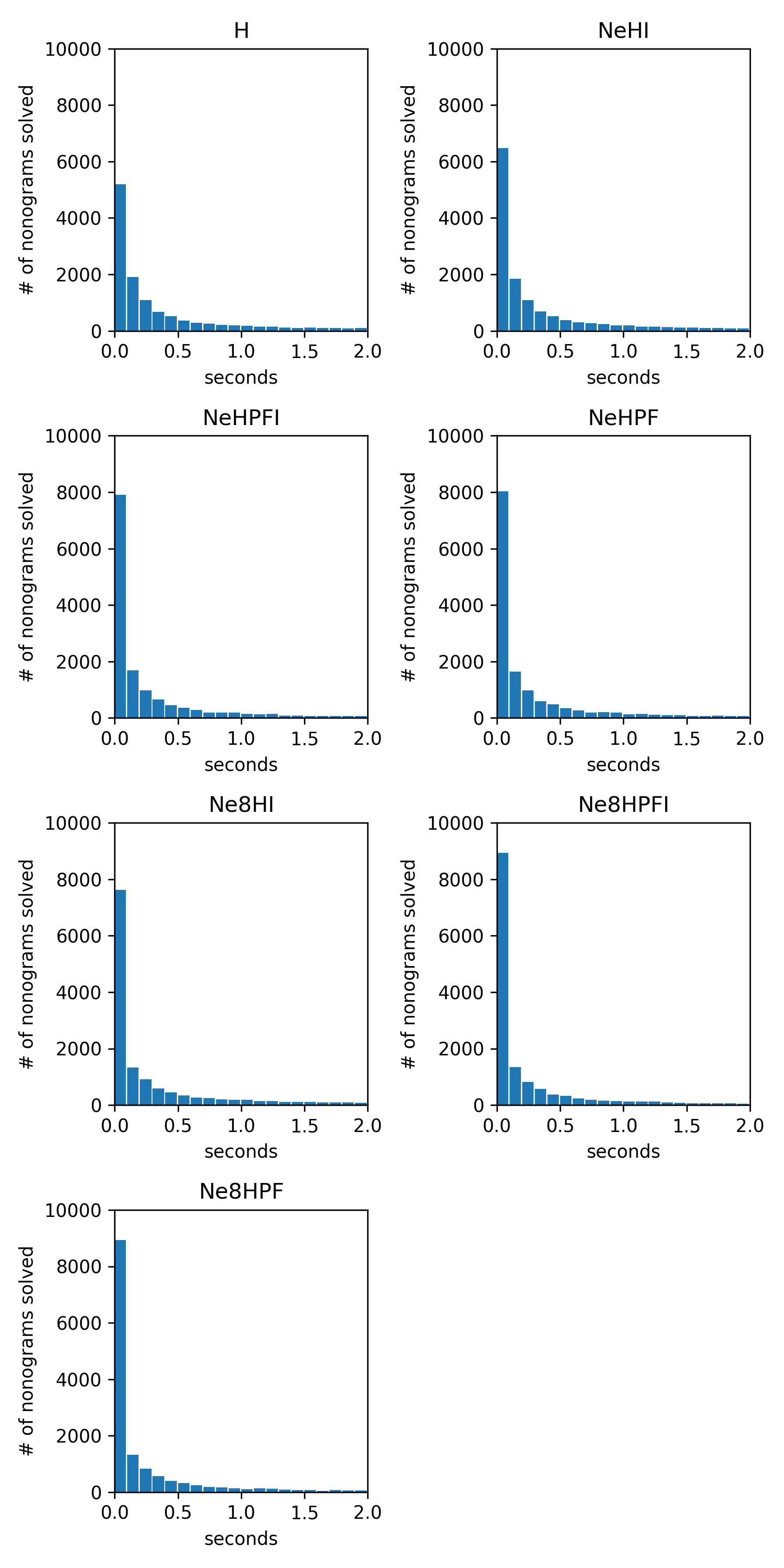}
	\end{center}
	\caption{Histograms depicting the mean time required to solve a board of size $10\times10$ using the different algorithms. We only show the first two seconds as these concentrated almost all information.}\label{fig:fig8}
\end{figure}

To analyze the time required for each variation of the heuristic algorithm in depth, Figure \ref{fig:fig8} depicts a histogram that describes the execution times of the algorithms during the first two seconds as they concentrated almost all information. Following the same concept, Table \ref{tab:table4}  presents the percentiles for each algorithm.  

It can be observed that the use of a neural network implied that the minimum time to solve a nonogram increased (from 0.001 to 0.0045 s) when we compare the solutions that used networks with the heuristic algorithm (H). However, this fact was irrelevant on more than $90\%$ of the boards (percentile 0.1). According to the Ne8Hi, Ne8HPF, and Ne8HPFI columns for the algorithms that used a network and the reflections, solving a board using the reflections reduced the time to solve a nonogram in $70\%$ of the cases (percentile 0.3). When we analyzed the time of the algorithms that used the partial erase technique, we observed that time improvements were achieved: comparing the NeHI and NeHPFI columns, starting at the 0.1 percentile, we obtained better results in 90\% of the boards, and when observing the results of Ne8HI and Ne8PFI, we obtained better results in 60\% of the boards. 

Finally, we determined that applying the intuition technique without using reflections implied a time penalty (see NeHPF and NeHPFI columns). However, using reflections (see Ne8HPF and Ne8HPFI columns) slightly improved the results in all cases, except in the worst case, where the use of intuition provoked a penalty of 4 seconds. 

\begin{table*}[htpb]
	
	\label{tab:table4}
	\begin{center}
		\begin{tabular}{p{0.2cm}p{1.1cm}p{1.1cm}p{1.2cm}p{1.2cm}p{1.3cm}p{1.3cm}p{1.4cm}}%
\textbf{P} & \textbf{H}      & \textbf{NeHi}   & \textbf{Ne8Hi}  & \textbf{NeHPF}  & \textbf{Ne8HPF} & \textbf{NeHPFI} & \textbf{Ne8HPFI} \\\hline
0.0                 & \textbf{0.0001} & 0.0045          & 0.0060          & 0.0043          & 0.0062          & 0.0045          & 0.0060           \\\hline
0.1                 & 0.0284          & 0.0050          & 0.0069          & \textbf{0.0046} & 0.0070          & \textbf{0.0049} & 0.0070           \\\hline
0.2                 & 0.0533          & \textbf{0.0052} & 0.0071          & \textbf{0.0047} & 0.0072          & \textbf{0.0050} & 0.0072           \\\hline
0.3                 & 0.0843          & 0.0334          & \textbf{0.0073} & 0.0182          & \textbf{0.0074} & 0.0189          & \textbf{0.0074}  \\\hline
0.4                 & 0.1299          & 0.0852          & 0.0230          & 0.0469          & 0.0139          & 0.0485          & 0.0137           \\\hline
0.5                 & 0.2042          & 0.1526          & 0.1003          & 0.0858          & 0.0520          & 0.0904          & 0.0519           \\\hline
0.6                 & 0.3298          & 0.2707          & 0.2187          & 0.1613          & 0.1123          & 0.1689          & 0.1123           \\\hline
0.7                 & 0.5895          & 0.5182          & 0.4545          & 0.3043          & 0.2504          & 0.3169          & 0.2492           \\\hline
0.8                 & 1.1843          & 1.0972          & 1.0360          & 0.6488          & 0.5551          & 0.6661          & 0.5529           \\\hline
0.9                 & 2.9784          & 2.7175          & 2.6103          & 1.8512          & 1.7439          & 1.8789          & 1.6834           \\\hline
1.0                 & 145.0626        & 151.3491        & 152.5022        & 52.0141         & 60.0945         & 66.7343         & 64.3024  \\\hline       
		\end{tabular}
  \caption{Percentiles of the execution time in seconds of the algorithms with boards of $10\times10$. The algorithms that contain an 8 in their name are the same algorithms as those in the line above, but they use reflections.}
	\end{center}
\end{table*}

The time to solve a nonogram did not follow a normal distribution in any of the proposed methods; therefore, we could not compare them by comparing their mean times. Thus, we analyzed the data using the Wilcoxon nonparametric test. To perform this test, we used the paired data Wilcoxon test with the difference in time between the two algorithms. We defined a p-value threshold of 0.05 to assume that the methods that we compared were significantly different. Table \ref{tab:table11} summarizes the results.  

As a first test, we compared the heuristic algorithm (H) and the algorithm with a network and partial and full erase using intuition (NeHPFI), and obtained a p-value near 0, so we could assume that the NeHPFI algorithm was superior. Next, we compared NeHPFI and NeHPF, similar to NeHPF, without using intuition, and we concluded that in this case (p-value near 0), the use of intuition led to the worst results. We can observe that in all comparisons between the algorithms using predictions with and without reflections, respectively (Ne8HI vs. NeHI, Ne8HPFI vs. NeHPF, and Ne8HPF vs. NeHPF), the use of this technique led to better results. Finally, we compared the Ne8HPFI and NeHPF algorithms, the two best among the previous comparisons. As the p-value was near 0, we concluded that Ne8HPFI was the best algorithm, as it solved nonograms in less time than the other algorithms.

\begin{table}[htpb]
	
	\label{tab:table11}
		\begin{center}
		\begin{tabular}{lll}%
			\textbf{Study} & \textbf{W} & \textbf{p-value}  \\ \hline
H vs \textbf{NeHPFI}       & 18461386   & $\approx$0  \\ \hline
NeHPFI vs \textbf{NeHPF}   & 98581813   & $\approx$0  \\ \hline
\textbf{Ne8HI} vs NeHI     & 37076704   & $\approx$0   \\ \hline
\textbf{Ne8HPFI} vs NeHPFI & 41639500  & $3.03e^{-206}$ \\ \hline
\textbf{Ne8HPF} vs NeHPF   & 53285564  & $1.09e^{-20}$  \\ \hline
\textbf{Ne8HPFI} vs Ne8HPF & 52985562  & $5.48e^{-23}$ \\ \hline
		\end{tabular}
  \caption{Wilcoxon contrast using paired data, with best algorithm in bold. Algorithms that contain an 8 in their name are the same as those in the line above, but they use reflections.}
	\end{center}
\end{table}

Regarding the results of boards of size $15\times15$, Table \ref{tab:table5} shows a summary of the execution of the 50 boards that we used as a test set with algorithms without reflections: H, NeHI, NeHPF, and NeHPFI. Table \ref{tab:table6} outlines the total execution time for all 50 boards and an estimation of the time required to solve all boards using the entire validation set. The execution of the four algorithms required 2 days, 16 hours, and 46 minutes, and if this time was extrapolated to the execution of the entire dataset, the execution would require 2 years, 94 days, and 6 hours.  

\begin{table}[htpb]
	
	\label{tab:table5}
	\begin{center}
		\begin{tabular}{lccc}
		
			\textbf{Algorithm} & \textbf{Mean}     & \textbf{Std}       & \textbf{Median} \\\hline
H                  & 1313.6846         & 6535.3968          & 5.4766          \\\hline
NeHI               & 1669.1414         & 7212.9861          & 5.4736          \\\hline
NeHPFI             & 1147.2674         & 5955.4429          & \textbf{2.8779} \\\hline
NeHPF              & \textbf{532.4148} & \textbf{1733.1702} & 3.0873         \\\hline\textbf{}
		\end{tabular}
  \caption{Statistical measurements of the execution time, in seconds, of the algorithms with boards of size $15\times15$ on 50 images of the test set.}
	\end{center}
\end{table}

\begin{table}[htpb]

	\label{tab:table6}
	\begin{center}
		\begin{tabular}{lll}%
			\textbf{Algorithm} & \textbf{Total}    & \textbf{Estimated} \\\hline
H                  & 18:14:45          & 232 days 05:40:20  \\\hline
NeHI               & 23:10:58          & 295 days 01:47:47  \\\hline
NeHPFI             & 15:56:04          & 202 days 19:36:03  \\\hline
NeHPF              & \textbf{07:23:41} & 94 days 02:55:04\\\hline
		\end{tabular}
	\end{center}
 \caption{Total execution time on 50 images of the test set and estimated execution time for the entire test set of the heuristic algorithms with boards of size $15\times15$.}
\end{table}

\subsection{Experiment 3: Genetic Algorithm} \label{sec:resultados:algs3}

We executed the genetic algorithm (GA) once on the test set of boards of size $10 \times 10$. The execution required 2 days, 18 hours, and 20 minutes, and it only managed to solve 15\% of the boards, without taking into account the boards that the network predicted correctly. Using the data obtained from this experiment, we can conclude that the GA yielded worse results than the heuristic algorithm, as it required a large amount of time to solve a nonogram and it did not have a guaranteed solution. This may be owing to the randomness of the algorithm or the fact that the \textit{fitness} function was not sufficient for the dimensionality of the solution space. 

\subsection{Results summary}
From the results obtained regarding the number of iterations and time required for each algorithm, we can conclude that even if the neural network cannot solve all boards, it can provide a good starting point to solve a nonogram that saves considerable computation time. The combination of the heuristic algorithm and neural networks outperformed the heuristic algorithm, DFS, and GA. 

\section{Conclusions} \label{sec:conclusion} 

In this study, we have presented an alternative to traditional algorithms designed to solve nonograms by adding neural networks to the decision process. Using the neural networks, it was possible to predict 27.25\% of boards of size $10 \times 10$ and 5.87\% of boards of size $15 \times 15$ correctly. 

The best solution was the combination of the heuristic algorithm with a neural network using reflections with partial and full erase, and intuition (Ne8HPFI). The GA exhibited the worst performance for the proposed task because it could not solve the entire set of boards. 

We also created a dataset \url{https://github.com/josebambu/NonoDataset}, which we have opened to the scientific community, to train and test the different networks and algorithms.
The project code is available at \url{https://gitlab.com/DLG-05/musi-tfm-nonograma}. 

\subsection{Further work}
Despite testing different fully connected network architectures to determine which was the best for predicting nonograms, we would like to evaluate other types of networks or other nonogram representations. 

Furthermore, we can design new \textit{fitness} functions to determine whether the conclusions that we obtained in our experiment with the GA are useful for improving its results. 

Likewise, for the variant of the algorithm that uses partial erase, a system could be proposed that would determine whether it is profitable to carry out partial deletion or to progress directly to complete deletion, thereby avoiding entering an invalid path and improving the overall system performance. 

\section*{Acknowledgments}
This work has a grant by project EXPLainable Artificial INtelligence systems for health and well-beING (EXPLAINING) (PID2019-104829RA-I00 / MCIN / AEI / 10.13039/501100011033).

\section*{Declarations}
\noindent \textbf{Conflict of Interests} The authors declare that they have no conflict of interest.

\noindent \textbf{Data and Code Availability} The datasets and the code generated during and/or used
during the current study are available in next url: \url{https://github.com/josebambu/NonoDataset}. 
The project code is available at \url{https://gitlab.com/DLG-05/musi-tfm-nonograma}

\bibliographystyle{elsarticle-num} 
\bibliography{biblio}

\end{document}